\documentclass[sigconf]{acmart}

\AtBeginDocument{%
  \providecommand\BibTeX{{%
    \normalfont B\kern-0.5em{\scshape i\kern-0.25em b}\kern-0.8em\TeX}}}

\copyrightyear{2022} 
\acmYear{2022} 
\setcopyright{acmcopyright}\acmConference[KDD '22]{Proceedings of the 28th ACM SIGKDD Conference on Knowledge Discovery and Data Mining}{August 14--18, 2022}{Washington, DC, USA}
\acmBooktitle{Proceedings of the 28th ACM SIGKDD Conference on Knowledge Discovery and Data Mining (KDD '22), August 14--18, 2022, Washington, DC, USA}
\acmPrice{15.00}
\acmDOI{10.1145/3534678.3539333}
\acmISBN{978-1-4503-9385-0/22/08}

\usepackage{times}
\usepackage{soul} 
\usepackage{url}
\usepackage[utf8]{inputenc}
\usepackage{graphicx}
\usepackage{amsmath}
\usepackage{booktabs}
\usepackage{subfig}
\usepackage{tabularx}
\usepackage{hyperref}
\usepackage{url}
\usepackage{graphicx}
\usepackage{makecell}
\usepackage{amsfonts}
\usepackage{amsmath,amsthm}
\usepackage{natbib}
\usepackage{multirow}
\usepackage{todonotes}
\usepackage{xcolor}
\usepackage{algorithm}
\usepackage{amsthm}
\usepackage{algpseudocode}
\usepackage{caption}
\usepackage{bm}
\usepackage{cuted}

\newtheorem{theorem}{Theorem}[]
\newtheorem{proposition}{Proposition}[]

\newtheorem{definition}{Definition}[]

\begin{document}

\title{Ultrahyperbolic Knowledge Graph Embeddings}


\author{Bo Xiong}
\orcid{0000-0002-5859-1961}
\affiliation{%
  \institution{University of Stuttgart}
  \city{Stuttgart}
  \country{Germany}
}
\email{bo.xiong@ipvs.uni-stuttgart.de}

\author{Shichao Zhu}
\affiliation{%
  \institution{IIE, Chinese Academy of Sciences}
  \institution{School of Cyber Security, UCAS}
  \city{Beijing}
  \country{China}
}
\email{zhushichao@iie.ac.cn}

\author{Mojtaba Nayyeri}
\affiliation{%
  \institution{University of Stuttgart}
  \city{Stuttgart}
  \country{Germany}
}

\author{Chengjin Xu}
\affiliation{%
  \institution{University of Bonn}
  \city{Bonn}
  \country{Germany}
}

\author{Shirui Pan}
\affiliation{%
  \institution{Monash University}
  \city{Melbourne}
  \country{Australia}
}

\author{Chuan Zhou}
\affiliation{%
  \institution{AMSS, Chinese Academy of Sciences}
  \city{Beijing}
  \country{China}
}

\author{Steffen Staab}
\affiliation{%
  \institution{University of Stuttgart}
  \institution{University of Southampton}
  \city{Stuttgart}
  \country{Germany}
}

\renewcommand{\shortauthors}{Xiong et al.}

\begin{abstract}
Recent knowledge graph (KG) embeddings have been advanced by hyperbolic geometry due to its superior capability for representing hierarchies. The topological structures of real-world KGs, however, are rather heterogeneous, i.e., a KG is composed of multiple distinct hierarchies and non-hierarchical graph structures. Therefore, a homogeneous (either Euclidean or hyperbolic) geometry is not sufficient for fairly representing such heterogeneous structures. To capture the topological heterogeneity of KGs, we present an ultrahyperbolic KG embedding (UltraE) in an ultrahyperbolic (or pseudo-Riemannian) manifold that seamlessly interleaves hyperbolic and spherical manifolds. In particular, we model each relation as a pseudo-orthogonal transformation that preserves the pseudo-Riemannian bilinear form. The pseudo-orthogonal transformation is decomposed into various operators (i.e., circular rotations, reflections and hyperbolic rotations), allowing for simultaneously modeling heterogeneous structures as well as complex relational patterns. Experimental results on three standard KGs show that UltraE outperforms previous Euclidean- and hyperbolic-based approaches.
\end{abstract}

\maketitle

\section{Introduction}
Knowledge graph (KG) embeddings, which map entities and relations into a low-dimensional space, have emerged as an effective way for a wide range of KG-based applications \citep{DBLP:journals/bmcbi/CelebiUYGDD19, DBLP:conf/www/0003W0HC19,DBLP:conf/wsdm/HuangZLL19}. In the last decade, various KG embedding methods have been proposed. Prominent examples include the \emph{additive} (or \emph{translational}) family \citep{DBLP:conf/nips/BordesUGWY13,DBLP:conf/aaai/WangZFC14,DBLP:conf/aaai/LinLSLZ15} and the \emph{multiplicative} (or \emph{bilinear}) family \citep{DBLP:conf/icml/NickelTK11,DBLP:journals/corr/YangYHGD14a,DBLP:conf/icml/LiuWY17}. 
Most of these approaches, however, are built on the Euclidean geometry that suffers from inherent limitations when dealing with hierarchical KGs such as WordNet \citep{DBLP:journals/cacm/Miller95}.
Recent studies \citep{DBLP:conf/nips/ChamiYRL19, DBLP:conf/nips/NickelK17} show that hyperbolic geometries (e.g., the Poincaré ball or Lorentz model) are more suitable for embedding hierarchical data because of their exponentially growing volumes.
Such \emph{tree-like} geometric space has been exploited in developing various hyperbolic KG embedding models such as MuRP \citep{DBLP:conf/nips/BalazevicAH19}, RotH \citep{DBLP:conf/acl/ChamiWJSRR20} and HyboNet \citep{DBLP:journals/corr/abs-2105-14686}, boosting the performance of link prediction on KGs with rich hierarchical structures and remarkably reducing the dimensionality.

\begin{figure}
    \centering
    \subfloat[\centering ]{{\includegraphics[width=.64\columnwidth]{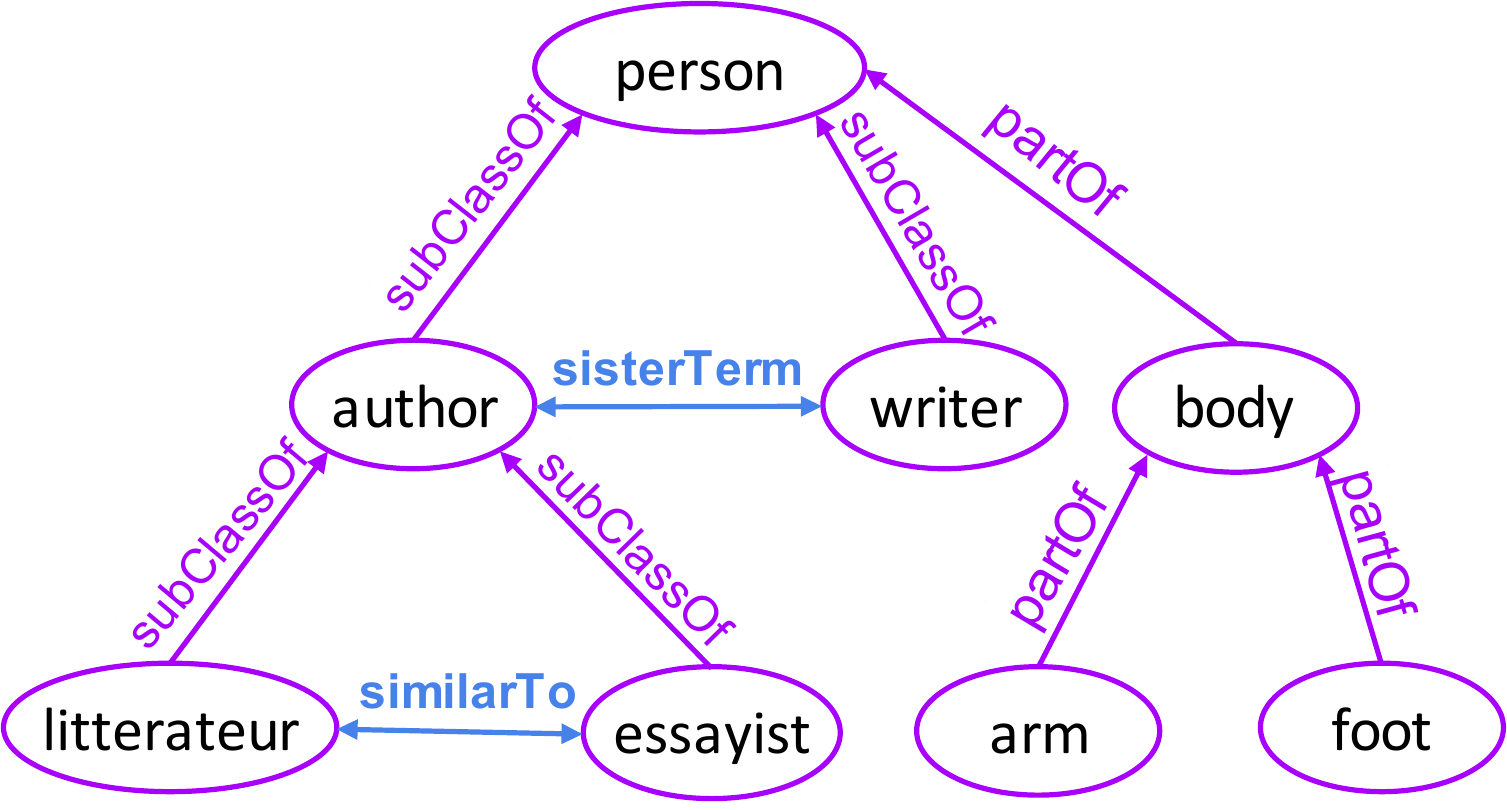}}}
    \subfloat[\centering ]{{\includegraphics[width=.36\columnwidth]{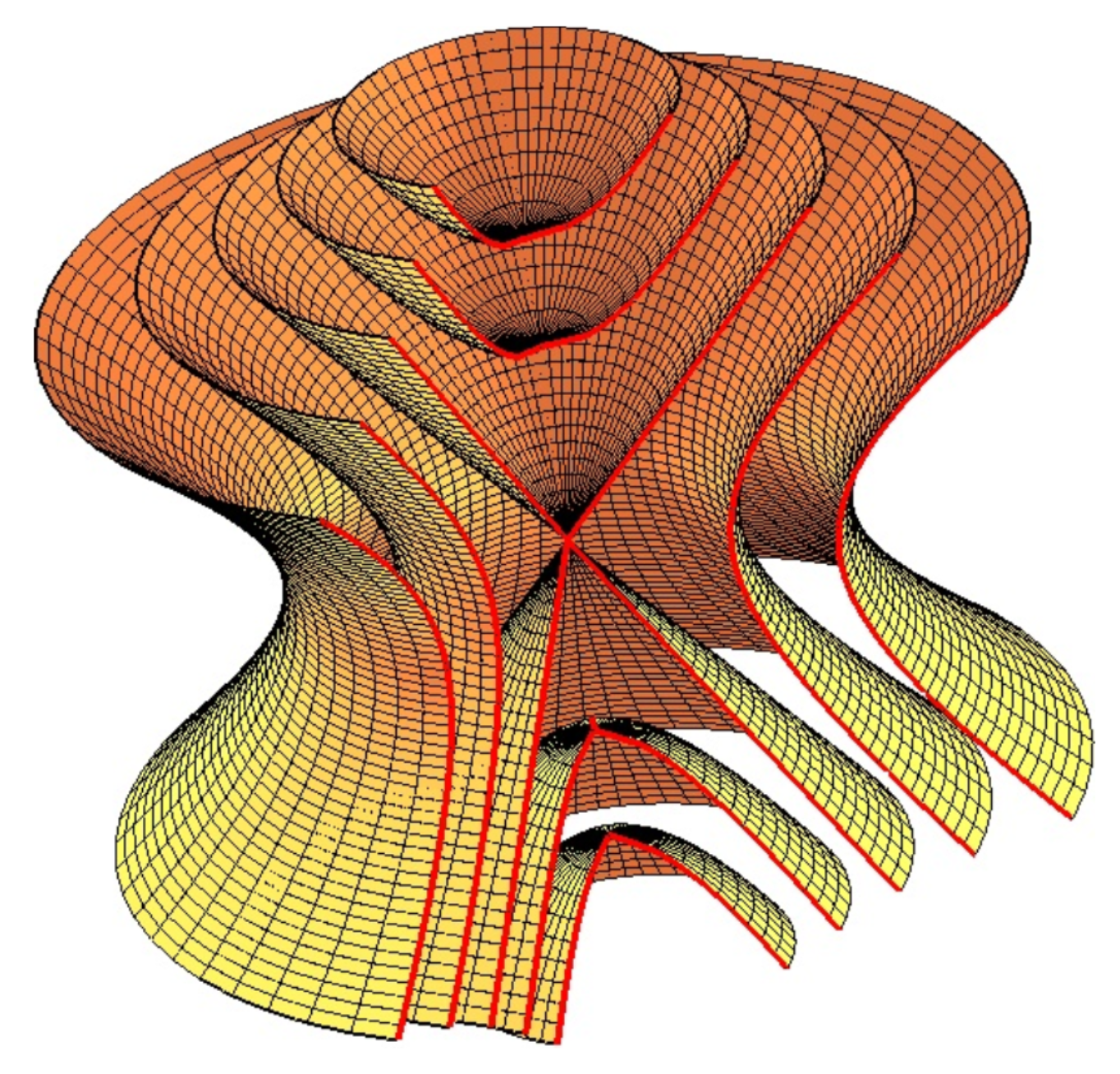} }}
    \caption{(a) A KG contains multiple distinct hierarchies (e.g., \textit{subClassOf} and \textit{partOf}) and non-hierarchical relations (e.g., \textit{similarTo} and \textit{sisterTerm}). 
    (b) An ultrahyperbolic manifold generalizing hyperbolic and spherical manifolds (figure from \citep{law2020ultrahyperbolic}).} 
    \label{fig:example}
    \vspace{-0.2cm}
\end{figure}

Although hierarchies are the most dominant structures, the real-world KGs usually exhibit heterogeneous topological structures, e.g., a KG consists of multiple hierarchical and non-hierarchical relations.
Typically, different hierarchical relations (e.g., \textit{subClassOf} and \textit{partOf}) form distinct hierarchies, while various non-hierarchical relations (e.g., \textit{similarTo} and \textit{sisterTerm}) capture the corresponding interactions between the entities at the same hierarchy level \citep{bai2021modeling}. 
Fig.\ref{fig:example}(a) shows an example of KG consisting of a heterogeneous graph structure. 
However, current hyperbolic KG embedding methods such as MuRP \citep{DBLP:conf/nips/BalazevicAH19} and HyboNet \citep{DBLP:journals/corr/abs-2105-14686} can only model a globally homogeneous hierarchy. 
RotH \citep{DBLP:conf/acl/ChamiWJSRR20} implicitly considers the topological "heterogeneity" of KGs and alleviates this issue by learning relation-specific curvatures that distinguish the topological characteristics of different relations. 
However, this does not entirely solve the problem, because hyperbolic geometry inherently \emph{mismatches} non-hierarchical data (e.g., data with cyclic structure) \citep{DBLP:conf/iclr/GuSGR19}.

To deal with data with heterogeneous topologies, a recent work \cite{DBLP:conf/www/WangWSWNAXYC21} learns KG embeddings in a product manifold and shows some improvements on KG completion. However, such product manifold is still a homogeneous space in which all data points have the same degree of heterogeneity (i.e., hierarchy and cyclicity), while KGs require relation-specific geometric mappings, e.g., relation \textit{partOf} should be more "hierarchical" than relation \textit{similarTo}.
Different from previous works, we consider an ultrahyperbolic manifold that seamlessly interleaves the hyperbolic and spherical manifolds. Fig.\ref{fig:example} (b) shows an example of ultrahyperbolic manifold that contains multiple distinct geometries. Ultrahyperbolic manifold has demonstrated impressive capability on embedding graphs with heterogeneous topologies such as hierarchical graphs with cycles \citep{law2020ultrahyperbolic,sim2021directed,DBLP:journals/corr/abs-2106-03134}.
However, such powerful representation space has not yet been exploited for embedding KGs with heterogeneous topologies.

In this paper, we propose ultrahyperbolic KG embeddings (UltraE), the first KG embedding method that simultaneously embeds multiple distinct hierarchical relations and non-hierarchical relations in a single but heterogeneous geometric space. 
The intuition behind the idea is that there exist multiple kinds of local geometries that could describe their corresponding  relations. For example, as shown in Fig.~\ref{fig:distance}(a), two points in the same \emph{circular conic section} are described by spherical geometry, while two points in the same half of a \emph{hyperbolic conic section} can be described by hyperbolic geometry.
In particular, we model entities as points in the ultrahyperbolic manifold and model relations as pseudo-orthogonal transformations, i.e., isometries in the ultrahyperbolic manifold. 
We exploit the theorem of hyperbolic Cosine-Sine decomposition \citep{DBLP:journals/siammax/StewartD05} to decompose the pseudo-orthogonal matrices into various geometric operations including circular rotations/reflections and hyperbolic rotations. Circular rotations/reflections allow for modeling relational patterns (e.g., composition), while hyperbolic rotations allow for modeling hierarchical graph structures. As Fig.~\ref{fig:distance}(b) shows, a combination of circular rotations/reflections and hyperbolic rotations induces various geometries including circular, elliptic, parabolic and hyperbolic geometries. 
These geometric operations are parameterized by Givens rotation/reflection \citep{DBLP:conf/acl/ChamiWJSRR20} and trigonometric functions, such that the number of relation parameters grows linearly w.r.t embedding dimensionality.
The entity embeddings are parametrized in Euclidean space and projected to the ultrahyperbolic manifold with differentiable and bijective mappings, allowing for stable optimization via standard Euclidean based gradient descent algorithms.

\noindent
\textbf{Contributions.} Our key contributions are summarized as follows:
\begin{itemize}
    \item We propose a novel KG embedding method, dubbed UltraE, that models entities in an ultrahyperbolic manifold seamlessly covering various geometries including hyperbolic, spherical and their combinations. UltraE enables modeling multiple hierarchical and non-hierarchical structures in a single but heterogeneous space. 
    \item We propose to decompose the relational transformation into various operators and parameterize them via Givens rotations/reflections such that the number of parameters is linear to the dimensionality. The decomposed operators allow for modeling multiple relational patterns including inversion, composition, symmetry, and anti-symmetry. 
    \item We propose a novel Manhattan-like distance in the ultrahyperbolic manifold, to retain the identity of indiscernibles while without suffering from the broken geodesic issues.
    \item We show the theoretical connection of UltraE with some existing approaches. Particularly, by exploiting the theorem of Lorentz transformation, we identify the connections between multiple hyperbolic KG embedding methods, including MuRP, RotH/RefH and HyboNet.
    \item We conduct extensive experiments on three standard benchmarks, and the experimental results show that UltraE outperforms previous Euclidean, hyperbolic and mixed-curvature (product manifold) baselines on  KG completion tasks. 
\end{itemize}

\section{Preliminaries}
In this paper, the points on a manifold are denoted by boldface lower letters $\mathbf{x,y}$. The matrices are denoted by boldface capital letters $\mathbf{U},\mathbf{V},\mathbf{I}$. 
The embedding spaces are denoted by blackboard bold capital letters like $\mathbb{R},\mathbb{H},\mathbb{S},\mathbb{U}$ denoting Euclidean, hyperbolic, spherical and ultrahyperbolic manifolds, respectively. 

\subsection{Pseudo-Riemannian Geometry} A pseudo-Riemannian manifold $(\mathcal{M},g)$ is a differentiable manifold $\mathcal{M}$ equipped with a metric tensor $g: T_{\mathbf{x}}\mathcal{M} \times T_{\mathbf{x}}\mathcal{M} \rightarrow \mathbb{R}$ defined in the entire tangent space $T_{\mathbf{x}}\mathcal{M}$, where $g$ is non-degenerate (i.e., $g(\mathbf{x}, \mathbf{y})=0$ for all $\mathbf{y} \in T_{\mathbf{x}}\mathcal{M} \backslash\{\mathbf{0}\}$ implies that $\mathbf{x}=\mathbf{0}$) and indefinite (i.e., $g$ could be positive, negative and zero). Such a metric is called a pseudo-Riemannian metric, defined as
\begin{small}
\begin{align}\label{eq:metric}
 \forall \mathbf{x},\mathbf{y} \in \mathbb{R}^{p,q}, \langle \mathbf{x}, \mathbf{y}\rangle_q = \sum_{i=1}^{p}\mathbf{x}_i \mathbf{y}_i-\sum_{j=p+1}^{p+q}\mathbf{x}_j\mathbf{y}_j, 
\end{align}
\end{small}
where $\mathbb{R}^{p,q}$ is a pseudo-Euclidean space (or space-time) with the dimensionality of $d=p+q$ where $p\geq 0, q\geq 0$. The space $\mathbb{R}^{p,q}$ has a rich background in physic \citep{o1983semi}. A point in $\mathbb{R}^{p,q}$ is interpreted as an \emph{event}, where the first $p$ dimensions and last $q$ dimensions are \emph{space-like} and \emph{time-like} dimensions, respectively. 
The pair $(p,q)$ is called the signature of space-time. 
Two important special cases are Riemannian ($q=0$) and Lorentz ($q=1$) geometrics that have positive definite metrics. The more general cases of pseudo-Riemannian geometry ($q \geq 2, p \geq 1$), however, do not need to possess positive definiteness, i.e., the scalar product induced by the metric could be positive (space-like), negative (time-like) or zero (light-like). 

\begin{figure}
    \centering
    \subfloat[\centering ]{{\includegraphics[width=.42\columnwidth]{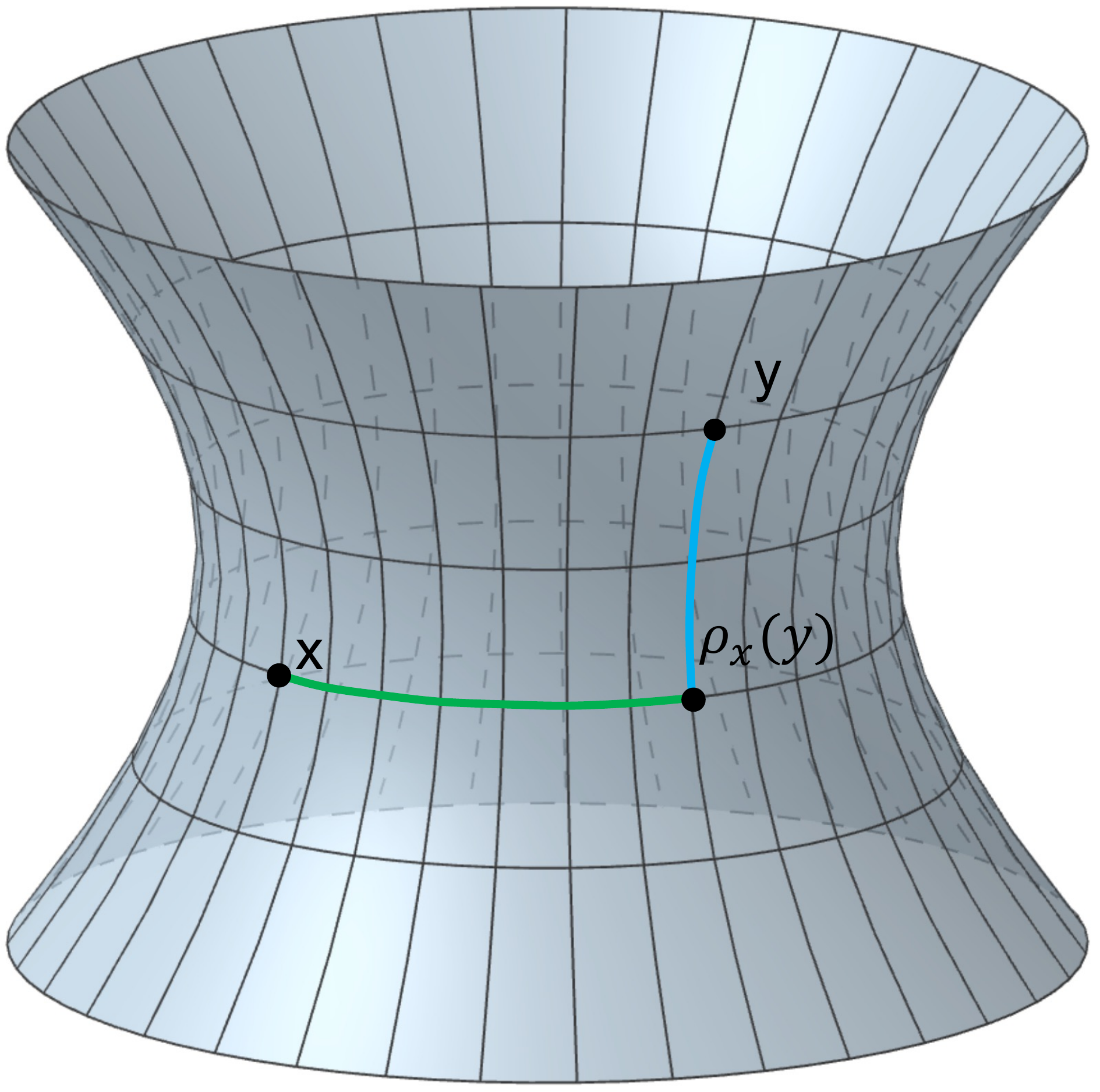} }}%
    \qquad
     \subfloat[\centering ]{{\includegraphics[width=0.48\columnwidth]{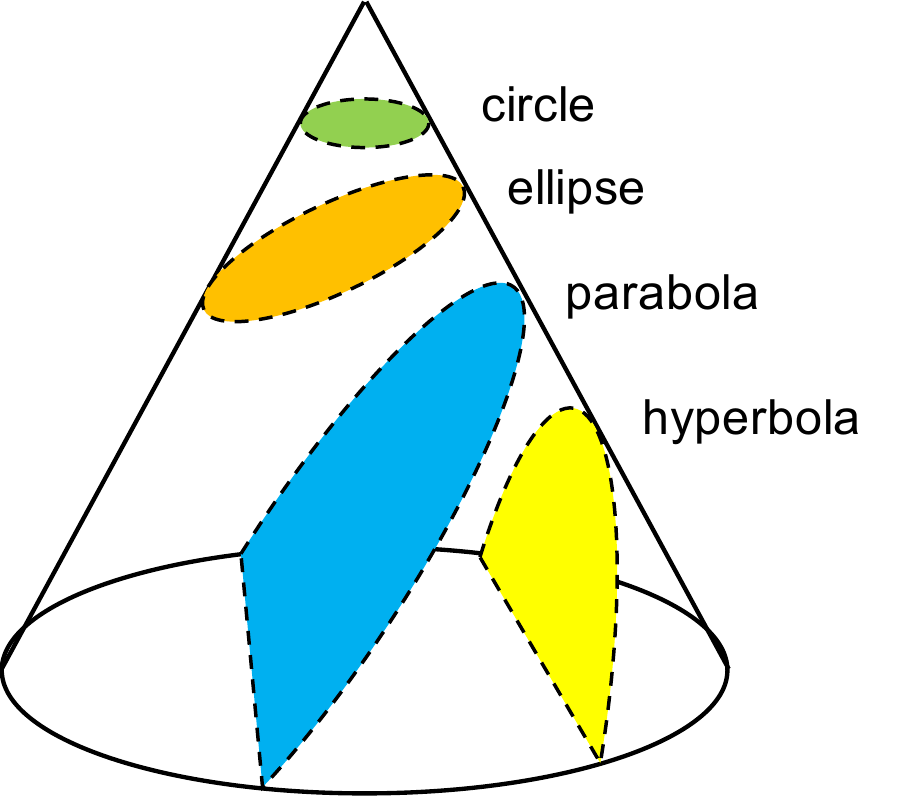}}}%
     \qquad
    \caption{(a) An illustration of a spherical (\textit{green}) geometry in the circular conic section and a hyperbolic (\textit{blue}) geometry in the hyperbolic conic section. The Manhattan-like distance of two points is defined by summing up the \emph{energy} moving from one point to another point with a circular rotation and a hyperbolic rotation. $\rho_{\mathbf{x}}(\mathbf{y})$ is a projection of $\mathbf{y}$ on an circular conic section crossing $\mathbf{x}$, such that $\rho_{\mathbf{x}}(\mathbf{y})$ and $\mathbf{x}$ are connected by a circular rotation while $\rho_{\mathbf{x}}(\mathbf{y})$ and $\mathbf{y}$ are connected by a hyperbolic rotation. (b) An illustration of the various geometries covered by the circular rotation and hyperbolic rotation, including circular, elliptic, parabolic and hyperbolic geometries.}
    \label{fig:distance}
\end{figure}

\subsection{Ultrahyperbolic Manifold} 
By exploiting the scalar product induced by Eq.~(\ref{eq:metric}), an ultrahyperbolic manifold (or pseudo-hyperboloid) is defined as a submanifold in the ambient space $\mathbb{R}^{p,q}$, given by

\begin{equation}\label{eq:pseudo-hyperboloid}
    \mathbb{U}_{\alpha}^{p, q}=\left\{\mathbf{x}=\left(x_{1}, x_{2}, \cdots, x_{p+q}\right)^{\top} \in \mathbb{R}^{p,q}:\|\mathbf{x}\|_{q}^{2}=-\alpha^{2}\right\},
\end{equation}
where $\alpha$ is a non-negative real number denoting the radius of curvature. $\|\mathbf{x}\|_{q}^{2}=\langle \mathbf{x}, \mathbf{x}\rangle_{q}$ is a norm of the induced scalar product. 

The ultrahyperbolic manifold $\mathbb{U}_{\alpha}^{p, q}$ can be seen as a generalization of hyperbolic and spherical manifolds, i.e., hyperbolic and spherical manifolds can be defined as special cases of ultrahyperbolic manifold by setting all time dimensions except one to be zero and setting all space dimensions to be zero, respectively, i.e., $\mathbb{H}_{\alpha} = \mathbb{U}_{\alpha}^{p,1}, \mathbb{S}_{\alpha} = \mathbb{U}_{\alpha}^{0,q}$. It is commonly known that hyperbolic and spherical manifolds are optimal geometric spaces for hierarchical and cyclic structures, respectively. Hence, ultrahyperbolic manifold is able to embed both hierarchical and cyclic structures. 

\section{Ultrahyperbolic Knowledge Graph Embeddings}
Let $\mathcal{E}$ and $\mathcal{R}$ denote the set of entities and relations.
A KG $\mathcal{K}$ consists of a set of triples $(h, r, t) \in \mathcal{K}$ where $h, t \in \mathcal{E}, r \in \mathcal{R}$ denote the head, the tail and their relation, respectively. 
The objective is to associate each entity with an embedding $\mathbf{e} \in \mathbb{U}^{p,q}$ in the ultrahyperbolic manifold, as well as a relation-specific transformation $f_r:\mathbb{U}^{p,q} \rightarrow \mathbb{U}^{p,q}$ that transforms one entity to another one in the ultrahyperbolic manifold. 


\subsection{Relation as Pseudo-Orthogonal Matrix}
We propose to model relations as pseudo-orthogonal (or $J$-orthogonal) transformations \citep{DBLP:journals/siamrev/Higham03}, a generalization of \emph{orthogonal transformation} in pseudo-Riemannian geometry. 
Formally, a real, square matrix $\mathbf{Q} \in \mathbb{R}^{d \times d}$ is called $J$-orthogonal if
\begin{equation}
    \mathbf{Q}^{T} \mathbf{J} \mathbf{Q}=\mathbf{J},
\end{equation}
where
$\mathbf{J}=\left[\begin{array}{cc} 
\mathbf{I}_{p} & \mathbf{0} \\
\mathbf{0} & -\mathbf{I}_{q}
\end{array}\right], p+q=d
$ and $\mathbf{I}_p$, $\mathbf{I}_q$ are identity matrices. 
$\mathbf{J}$ is called a signature matrix of signature $(p,q)$. Such $J$-orthogonal matrices form a multiplicative group called pseudo-orthogonal group $O(p,q)$. Conceptually, a matrix $\mathbf{Q} \in O(p,q)$ is an \emph{isometry} (distance-preserving transformation) in the ultrahyperbolic manifold that preserves the bilinear form (i.e., $\forall \mathbf{x} \in \mathbb{U}^{p,q}, \mathbf{Q}\mathbf{x} \in \mathbb{U}^{p,q} $). Therefore, the matrix acts as a linear transformation in the ultrahyperbolic manifold.

There are two challenges to model relations as $J$-orthogonal transformations: 1) $J$-orthogonal matrix requires $\mathcal{O}(d^2)$ parameters. 2) Directly optimizing the $J$-orthogonal matrices results in constrained optimization, which is practically challenging within the standard gradient based framework. 

\subsubsection{Hyperbolic Cosine-Sine Decomposition}
To solve these issues, we seek to decompose the $J$-orthogonal matrix by exploiting the Hyperbolic Cosine-Sine (CS) Decomposition. 
\begin{proposition}[Hyperbolic CS Decomposition \citep{DBLP:journals/siammax/StewartD05}]\label{prop:cs_decomposition}
Let $\mathbf{Q}$ be $J$-orthogonal and assume that $q \leq p.$ Then there are orthogonal matrices $\mathbf{U}_{1}, \mathbf{V}_{1} \in \mathbb{R}^{p \times p}$ and $\mathbf{U}_{2}, \mathbf{V}_{2} \in \mathbb{R}^{q \times q}$ s.t.
\begin{equation}\label{eq:cs_decomposition}
\mathbf{Q}=\left[\begin{array}{cc}
\mathbf{U}_{1} & \mathbf{0} \\
\mathbf{0} & \mathbf{U}_{2}
\end{array}\right]\left[\begin{array}{ccc}
\mathbf{C} & \mathbf{0} & \mathbf{S} \\
\mathbf{0} & I_{p-q} & \mathbf{0} \\
\mathbf{S} & \mathbf{0} & \mathbf{C} 
\end{array}\right]\left[\begin{array}{cc}
\mathbf{V}_{1}^{T} & \mathbf{0} \\
\mathbf{0} & \mathbf{V}_{2}^{T}
\end{array}\right],
\end{equation}
where $\mathbf{C}=\operatorname{diag}\left(c_1, \ldots, c_q\right), \mathbf{S}=\operatorname{diag}\left(s_1, \ldots, s_q\right)$ and $\mathbf{C}^{2}-\mathbf{S}^{2}=\mathbf{I}_q$. For cases where $q > p$, the decomposition can be defined analogously. For simplicity, we only consider $q \leq p$.
\end{proposition}

\begin{figure}[]
    \centering
    \includegraphics[width=0.8\columnwidth]{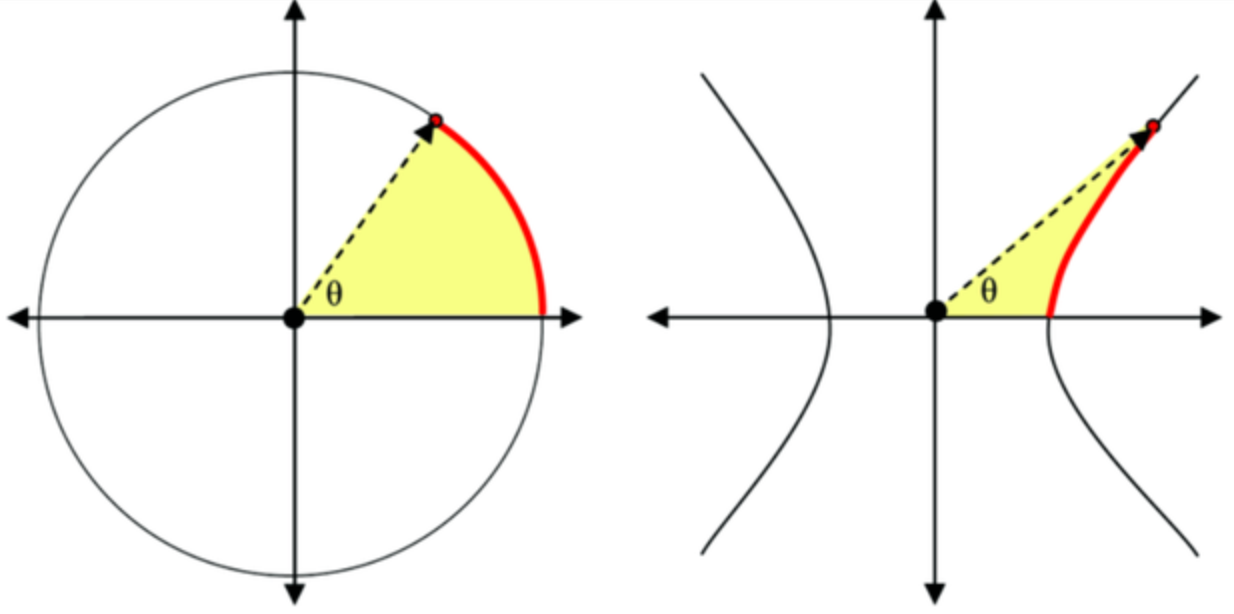}
    \caption{A two-dimensional example of circular rotation (\textit{left}) and hyperbolic rotation (\textit{right}), where $\theta$ is the angle of rotations.}
    \label{fig:rotation}
\end{figure}

Geometrically, the $J$-orthogonal matrix is decomposed into various geometric operators. The orthogonal matrices $\mathbf{U}_{1}, \mathbf{V}_{1}$ represent circular rotation or reflection (depending on their determinant) \footnote{Depending on the determinant, a orthogonal matrix $\mathbf{U}$ denotes a rotation iff $\operatorname{det}(\mathbf{U})=1$ or a reflection iff $\operatorname{det}(\mathbf{U})=-1$} in the space dimension, while $\mathbf{U}_{2}, \mathbf{V}_{2}$ represent circular rotation or reflection in the time dimension. The intermediate matrix that is uniquely determined by $\mathbf{C},\mathbf{S}$, denotes a hyperbolic rotation (analogous to the "circular rotation") across the space and time dimensions. Fig. \ref{fig:rotation} shows a $2$-dimensional example of circular rotation and hyperbolic rotation. 

It is worth noting that both circular rotation/reflection and hyperbolic rotation are important operations for KG embeddings. 
On the one hand, circular rotations/reflections are able to model complex relational patterns including inversion, composition, symmetry, or anti-symmetry. Besides, these relational patterns usually form
some non-hierarchies (e.g., cycles). Hence, circular rotations/reflections inherently encode non-hierarchical graph structures. 
Hyperbolic rotation, on the other hand, is able to model hierarchies, i.e., by connecting entities at different levels of hierarchies. 
Therefore, the decomposition in proposition \ref{prop:cs_decomposition} shows that 
$J$-orthogonal transformation is powerful for representing both relational patterns and graph structures.

\subsection{Relation Parameterization}
In light of this, we now parameterize circular rotation/reflection and hyperbolic rotation, respectively.
\subsubsection{Circular Rotation/Reflection.} Parameterizing circular rotation or reflection via orthogonal matrices is non-trivial and there are some \emph{trivialization} approaches such as using Cayley Transform \citep{shepard2015representation}. However, such parameterization requires  $\mathcal{O}(d^2)$ parameter complexity. To simplify the complexity, we consider Given transformations denoted by $2 \times 2$ matrices. Suppose the number of dimension $p,q$ are even, circular rotation and reflection can be denoted by block-diagonal matrices of the form, given as
\begin{equation}\label{eq:rot_ref}
    \begin{array}{c}
    \operatorname{\mathbf{Rot}}\left(\Theta_{r}\right)=\operatorname{diag}\left(\mathbf{G}^{+}\left(\theta_{r, 1}\right), \ldots, \mathbf{G}^{+}\left(\theta_{r, \frac{p+q}{2}}\right)\right) \\
    \operatorname{\mathbf{Ref}}\left(\Phi_{r}\right)=\operatorname{diag}\left(\mathbf{G}^{-}\left(\phi_{r, 1}\right), \ldots, \mathbf{G}^{-}\left(\phi_{r, \frac{p+q}{2}}\right)\right) \\
    \text {where} \quad \mathbf{G}^{\pm}(\theta):=\left[\begin{array}{cc}
    \cos (\theta) & \mp \sin (\theta) \\
    \sin (\theta) & \pm \cos (\theta)
    \end{array}\right]
    \end{array},
\end{equation}
where $\Theta_{r}:=\left(\theta_{r, i}\right)_{i \in\left\{1, \ldots \frac{p+q}{2}\right\}}$ and $\Phi_{r}:=\left(\phi_{r, i}\right)_{i \in\left\{1, \ldots \frac{p+q}{2}\right\}}$ are relation-specific parameters. 

Although circular rotation is theoretically able to infer symmetric patterns \citep{DBLP:conf/emnlp/WangLLS21} (i.e., by setting rotation angle $\theta=\pi$ or $\theta=0$), circular reflection can more effectively represent symmetric relations since their second power is the identity. AttH \citep{DBLP:conf/emnlp/WangLLS21} combines circular rotations and circular reflections by using an attention mechanism learned in the tangent space, which requires additional parameters.
We also combine circular rotation and circular reflection operators but in a different way. Since the $J$-orthogonal matrix is decomposed into two rotation/reflection matrices, we set the first matrix in Eq.~(\ref{eq:cs_decomposition}) to be circular rotation while the third part to be circular reflection matrices, given by
\begin{equation}\small\label{eq:uv}
\begin{split}
\mathbf{U_{\Theta_r}}=\left[\begin{array}{cc}
\operatorname{\operatorname{\mathbf{Rot}}}\left(\Theta_{r_{p}}\right) & \mathbf{0} \\
\mathbf{0} & \operatorname{\operatorname{\mathbf{Rot}}}\left(\Theta_{r_{q}}\right)
\end{array}\right],
\end{split}
\quad
\begin{split}
\mathbf{V_{\Phi_r}}=\left[\begin{array}{cc}
\operatorname{\operatorname{\mathbf{Ref}}}\left(\Phi_{r_{p}}\right) & \mathbf{0} \\
\mathbf{0} & \operatorname{\operatorname{\mathbf{Ref}}}\left(\Phi_{r_{q}}\right)
\end{array}\right]
\end{split},
\end{equation}

Clearly, the parameterization of circular rotation and reflection in Eq.~ (\ref{eq:rot_ref}), as well as the combination of them in Eq.(\ref{eq:uv}), lose a certain degree of freedom of $J$-orthogonal transformation. However, it 1) results in a linear ($\mathcal{O}(d)$) memory complexity of relational embeddings; 2) significantly reduces the risk of overfitting; and 3) is sufficiently expressive to model complex relational patterns as well as graph structures. This is similar to many other Euclidean models, such as SimplE \citep{DBLP:conf/nips/Kazemi018}, that sacrifice some degree of freedoms of the multiplicative model (i.e., RESCAL \citep{DBLP:conf/icml/NickelTK11}) parameterized by quadratic matrices while pursuing a linearly complex, less overfitting and highly expressive relational model. Our ablation studies will show that such combination of rotation and reflection results in performance gains.

\subsubsection{Hyperbolic Rotation.} 
The hyperbolic rotation matrix is parameterized by two diagonal matrices $\mathbf{C},\mathbf{S}$ that satisfy the condition $\mathbf{C}^{2}-\mathbf{S}^{2}=\mathbf{I}$. The hyperbolic rotation matrix can be seen as a generalization of the $2\times2$ hyperbolic rotation given by $\left[\begin{array}{ll}\cosh (\mu) & \sinh (\mu) \\ \sinh (\mu) & \cosh (\mu)\end{array}\right]$, where the trigonometric functions $\sinh$ and $\cosh$ are hyperbolic versions of the $\sin$ and $\cos$ functions. Clearly, it satisfies the condition $\cosh (\mu)^2-\sinh(\mu)^2=1$. Analogously, to satisfy the condition $\mathbf{C}^{2}-\mathbf{S}^{2}=\mathbf{I}$, we parameterize $\mathbf{C},\mathbf{S}$ by diagonal matrices
\begin{align}
    & \mathbf{C}(\mu)=\operatorname{diag}\left(\cosh (\mu_1), \ldots, \cosh (\mu_q)\right), \\
    & \mathbf{S}(\mu)=\operatorname{diag}\left(\sinh (\mu_1), \ldots, \sinh (\mu_q)\right),
\end{align}
where $\mu=(\mu_1, \cdots, \mu_q)$ is the parameter of hyperbolic rotation to learn. 
Therefore, the hyperbolic rotation matrix can be denoted by
\begin{equation}\scriptsize\label{eq:translation}
\mathbf{H}_{\mu_r}=\left[\begin{array}{ccc}
\operatorname{diag}\left(\cosh (\mu_{r,1}), \ldots, \cosh (\mu_{r,q})\right)  & \mathbf{0} & \operatorname{diag}\left(\sinh (\mu_{r,1}), \ldots, \sinh (\mu_{r,q})\right) \\
\mathbf{0} & I_{p-q} & \mathbf{0}, \\
\operatorname{diag}\left(\sinh (\mu_{r,1}), \ldots, \sinh (\mu_{r,q})\right) & \mathbf{0} & \operatorname{diag}\left(\cosh (\mu_{r,1}), \ldots, \cosh (\mu_{r,q})\right) 
\end{array}\right],
\end{equation}

Given the parameterization of each component, the final transformation function of relation $r$ is given by
\begin{equation}\label{eq:f_r}
    f_r=\mathbf{U}_{\theta_r} \mathbf{H}_{\mu_r} \mathbf{V}_{\Phi_r}.
\end{equation}
Notably, the combination of circular rotation/reflection and hyperbolic rotation covers various kinds of geometric transformations in the ultrahyperbolic manifold, including circular, elliptic, parabolic, and hyperbolic transformations (See Fig.~\ref{fig:distance}(a)). Hence, our relational embedding is able to work with all corresponding geometrical spaces.


\subsection{Objective and Manhattan-like Distance}

\subsubsection{Objective Function.} 
Given $f_r$ and entity embeddings $e$, we design a score function for each triplet $(h,r,t)$ as
\begin{equation}
s(h, r, t)=-d_{\mathbf{U}}^{2}\left(f_{r}\left(\mathbf{e}_{h}\right), \mathbf{e}_{t}\right)+b_{h}+b_{t}+\delta,
\end{equation}
where $f_{r}\left(\mathbf{e}_{h}\right) = \mathbf{U}_{\theta_r} \mathbf{H}_{\mathbf{\mu}_r} \mathbf{V}_{\Phi_r} \mathbf{e_h}$, and $\mathbf{e}_{h}, \mathbf{e}_{t} \in \mathbb{U}^{p,q}$ are the embeddings of head entity $h$ and tail entity $t$, $b_h,b_t \in \mathbf{R}^d$ are entity-specific biases, and each bias defines an entity-specific \emph{sphere of influence} \citep{DBLP:conf/nips/BalazevicAH19} surrounding the center point. $\delta$ is a global margin hyper-parameter. $d_{\mathbb{U}}(\cdot)$ is a function that quantifies the nearness/distance between two points in the ultrahyperbolic manifold. 

\subsubsection{Manhattan-like Distance.} 
Defining a proper distance $d_{\mathbb{U}}(\cdot)$ in the ultrahyperbolic manifold is non-trivial. Different from Riemannian manifolds that are geodesically connected, ultrahyperbolic manifolds are not geodesically connected, and there exist \emph{broken cases} in which the geodesic distance is not defined \citep{law2020ultrahyperbolic}. 
Some approximation approaches \citep{law2020ultrahyperbolic,DBLP:journals/corr/abs-2106-03134} are proposed and satisfy some of the axioms of a classic metric (e.g., symmetric premetric). However, these distances suffer from the lack of the identity of indiscernibles, that is, one may have $\mathbf{d}_{\mathbb{U}}(\mathbf{x}, \mathbf{y})=0$ for some distinct points $\mathbf{x}\neq\mathbf{y}$. This is not a problem for metric learning that learns to preserve the pair-wise distances
\citep{law2020ultrahyperbolic,DBLP:journals/corr/abs-2106-03134}. However, our preliminary experiments find that the geodesic distance lacks the identity of indiscernibles results in unstable and non-convergent training. We conjecture this is due to the fact that the target of KG embedding is different from the graph embedding aiming at preserving pair-wise distance. KG embedding aims at satisfying $f_r(\mathbf{e}_h) \approx \mathbf{e}_t$ for each positive triple $(h,r,t)$ while not for negative triples. Hence, we need to retain the \emph{identity of indiscernibles}, that is, $\mathbf{d}_{\mathbb{U}}(\mathbf{x}, \mathbf{y})=0 \Leftrightarrow \mathbf{x}=\mathbf{y}$.


To address this issue, we propose a novel Manhattan-like distance function, which is defined by a composition of a spherical distance and a hyperbolic distance. 
Fig.~\ref{fig:distance} shows the Manhattan-like distance. 
Formally, given two points $\mathbf{x},\mathbf{y} \in \mathbb{U}^{p,q}$, we first define a projection $\rho_{\mathbf{x}}(\mathbf{y})$ of $\mathbf{y}$ on the \emph{circular conic section} crossing $\mathbf{x}$, such that $\rho_{\mathbf{x}}(\mathbf{y})$ and $\mathbf{x}$ share the same space dimension while $\rho_{\mathbf{x}}(\mathbf{y})$ and $y$ lie on a hyperbolic subspace. 
\begin{equation}\label{eq:dis_proj}
    \rho_{\mathbf{x}}(\mathbf{y}) = \left(\begin{array}{c}
    \mathbf{ \mathbf{x}_p } \\
    \alpha \mathbf{y}_q \frac{\|x_p\|}{\|y_p\|} 
    \end{array}\right),
\end{equation}
This projection makes that $\mathbf{x}$ and $\rho_\mathbf{x}(\mathbf{y})$ are connected by a purely space-like geodesic while $\rho_\mathbf{x}(\mathbf{y}),\mathbf{y}$ are connected by a purely time-like geodesic. 
The distance function of $\mathbb{U}$ hence can be defined as a Manhattan-like distance, i.e., the sum of the two distances, given as
\begin{align}\label{eq:distance}
     d_{\mathbb{U}}(\mathbf{x},\mathbf{y}) = \min\{& d_\mathbb{S}(\mathbf{y},\rho_\mathbf{y}(\mathbf{x})) + d_\mathbb{H}(\rho_\mathbf{y}(\mathbf{x}),\mathbf{x}), \\ 
     & d_\mathbb{S}(\mathbf{x},\rho_\mathbf{x}(\mathbf{y})) + d_\mathbb{H}(\rho_\mathbf{x}(\mathbf{y}),\mathbf{y})\},
\end{align}
where $d_\mathbb{S}$ and $d_\mathbb{H}$ are spherical and hyperbolic distances, respectively, which are well-defined and maintain the identity of indiscernibles.

\subsection{Optimization}
For each triplet $(h,r,t)$, we create $k$ negative samples by randomly corrupting its head or tail entity. The probability of a triple is calculated as $p=\sigma(s(h, r, t))$ where $\sigma(.)$ is a sigmoid function. We minimize the binary cross entropy loss, given as
\begin{equation}
    \mathcal{L}=-\frac{1}{N} \sum_{i=1}^{N}\left(\log p^{(i)}+\sum_{j=1}^{k} \log \left(1-\tilde{p}^{(i, j)}\right)\right),
\end{equation}
where $p^{(i)}$ and $\tilde{p}^{(i, j)}$ are the probabilities for positive and negative triplets respectively, and $N$ is the number of samples.

Notably, directly optimizing the embeddings in ultrahyperbolic manifold is challenging. The issue is caused by the fact that there exist some points that cannot be connected by a geodesic in the manifold (hence no tangent direction for gradient descent). One way to sidestep the problem is to define entity embeddings in the Euclidean space and use a \emph{diffeomorphism} to map the points into the manifold. 
In particular, we consider the following diffeomorphism. 
\begin{theorem}\label{lm:sbr}[Diffeomorphism \citep{DBLP:journals/corr/abs-2106-03134}]
For any point $\mathbf{x} \in \mathbb{U}_{\alpha}^{p, q}$, there exists a diffeomorphism $\psi: \mathbb{U}_{\alpha}^{p, q} \rightarrow \mathbb{R}^{p} \times \mathbb{S}_{\alpha}^{q}$ that maps $\mathbf{x}$ into the product manifolds of a sphere and the Euclidean space. The mapping and its inverse are given by
\end{theorem}
\begin{equation}
    \psi(\mathbf{x})=\left(\begin{array}{c}
    \mathbf{s} \\
    \alpha \frac{\mathbf{t}}{\|\mathbf{t}\|}
    \end{array}\right), \quad \psi^{-1}(\mathbf{z})=\left(\begin{array}{c}
    \mathbf{v}\\
    \frac{\sqrt{\alpha^2+\|\mathbf{v}\|^{2}}}{\alpha} \mathbf{u}
    \end{array}\right) \text {, }
\end{equation}
where $\mathbf{x}=\left(\begin{array}{c}\mathbf{s} \\ \mathbf{t}\end{array}\right) \in \mathbb{U}_{\alpha}^{p,q}$ with $\mathbf{s} \in \mathbb{R}^{p}$ and $\mathbf{t} \in \mathbb{R}_{*}^{q}$. $\mathbf{z}=\left(\begin{array}{c}\mathbf{v} \\ \mathbf{u} \end{array}\right) \in \mathbb{R}^{p} \times \mathbb{S}_{\alpha}^{q} $ with $\mathbf{v} \in \mathbb{R}^{p}$ and $\mathbf{u} \in \mathbb{S}_{\alpha}^{q}$. 

With these mappings, any vector $\mathbf{x} \in \mathbb{R}^{p} \times \mathbb{R}_{*}^{q}$ can be mapped to $\mathbb{U}_{\alpha}^{p, q}$ by a double projection $\varphi=\psi^{-1} \circ \psi$. Note that since the diffeomorphism is differential and bijective, the canonical chain rule can be exploited to perform standard gradient descent optimization. 

\section{Theoretical Analysis}
In this section, we provide some theoretical analyses of UltraE and some related approaches.

\subsection{Complexity Analysis}
To make the model scalable to the size of the current KGs and keep up with their growth, a KG embedding model should have linear time and parameter (space) complexity \citep{DBLP:journals/corr/abs-1304-7158,DBLP:conf/nips/Kazemi018}. 
In our case, the number of relation parameters of circular rotation, circular reflection, and hyperbolic rotation grows linearly with the dimensionality given by $p+q$. The total number of parameters is then $\mathcal{O}( (N_e + N_r)\times d)$, where $N_e$ and $N_r$ are the numbers of entities and relations and $d=p+q$ is the embedding dimensionality. Similar to TransE \cite{DBLP:conf/nips/BordesUGWY13} and RotH \cite{DBLP:conf/acl/ChamiWJSRR20}, UltraE has time complexity $\mathcal{O}(d)$. 

\subsection{Connections with Hyperbolic Methods}
UltraE has close connections with some existing hyperbolic KG embedding methods, including HyboNet \citep{DBLP:journals/corr/abs-2105-14686}, RotH/RefH \citep{DBLP:conf/acl/ChamiWJSRR20}, and MuRP \citep{DBLP:conf/nips/BalazevicAH19}. To show this, we first introduce Lorentz transformation. 
\begin{definition}\label{def:lorent}
Lorentz transformation is a pseudo-orthogonal transformation with signature $(p,1)$.
\end{definition}

\noindent
\textbf{HyboNet} \citep{DBLP:journals/corr/abs-2105-14686} embeds entities as points in a Lorentz space and models relations as Lorentz transformations. According to Definition \ref{def:lorent}, we have the following proposition.

\begin{proposition}
UltraE, if parameterized by a full $J$-orthogonal matrix, generalizes HyboNet. 
\end{proposition}
That is, HyboNet is the case of UltraE (with full $J$-orthogonal matrix parameterization) where $q=1$.

By exploiting the polar decomposition \citep{ratcliffe1994foundations}, a Lorentz transformation matrix $\mathbf{T}$ can be decomposed into $\mathbf{T}=\mathbf{R}_{\mathbf{U}} \mathbf{R}_{\mathbf{b}}$, where
\begin{equation}
    \mathbf{R_{U}}=\left[\begin{array}{cc}
\mathbf{U} & \mathbf{0} \\
\mathbf{0} & 1
\end{array}\right], \mathbf{R_{b}}=\left[\begin{array}{cc}
\left(\mathbf{I}+\mathbf{b} \mathbf{b}^{\top}\right)^{\frac{1}{2}} & \mathbf{b}^{\top} \\
\mathbf{b} & \sqrt{1+\|\mathbf{b}\|_{2}^{2}}
\end{array}\right],
\end{equation}
where $\mathbf{R_{U}}$ is an orthogonal matrix. 
In Lorentz geometry, $\mathbf{R_{U}}$ and $\mathbf{R_{b}}$ are called Lorentz rotation and Lorentz boost, respectively. $\mathbf{R_{U}}$ represents rotation or reflection in space dimension (without changing the time dimension), while $\mathbf{R_{b}}$ denotes a hyperbolic rotation across the time dimension and each space dimension. 
\cite{DBLP:journals/spl/TabaghiD21} established an equivalence between Lorentz boost and Möbius addition (or hyperbolic translation). Hence, HyboNet inherently models each relation as a combination of a rotation/reflection and a hyperbolic translation. 

\noindent
\textbf{RotH/RefH} \citep{DBLP:conf/acl/ChamiWJSRR20}, interestingly, also models each relation as a combination of a rotation/reflection and a hyperbolic translation that is implemented by Möbius addition. 
Hence, HyboNet subsumes RotH/RefH,\footnote{Note that RotH/RefH consider Poincaré Ball while HyboNet considers Lorentz model. The subsumption still holds since Poincaré Ball is isometric to the Lorentz model.} where the equivalence cannot hold because the rotation/reflection of RotH/RefH is parameterized by the Givens rotation/reflection \citep{DBLP:conf/acl/ChamiWJSRR20}. 

\noindent
\textbf{MuRP} \citep{DBLP:conf/nips/BalazevicAH19} models relations as a combination of Möbius multiplication (with diagonal matrix) and Möbius addition. Note that \citep{DBLP:journals/corr/abs-2105-14686} established a fact that a Lorentz rotation is equivalent to Möbius multiplication, and \cite{DBLP:journals/spl/TabaghiD21} proved that Lorentz boost is equivalent to Möbius addition. Hence, HyboNet subsumes MuRP, where the equivalence cannot hold because the Möbius multiplication in MuRP is parameterized by a diagonal matrix.

To sum up, UltraE generalizes HyboNet to allow for arbitrary signature $(p,q)$, while HyboNet subsumes RotH/RefH and MuRP. 



\begin{table}[]
    \centering
     \caption{The statistics of KGs, where $\xi_{G}$ measures the tree-likeness (the lower the $\xi_{G}$ is, the more tree-like the KG is).}
    \begin{tabular}{lcccc}
        \hline Dataset & \#entities & \#relations & \#triples & $\xi_{G}$ \\
        \hline WN18RR & $41 \mathbf{k}$ & 11 & $93 \mathbf{k}$ & $-2.54$ \\
        \hline FB15k-237 & $15 \mathbf{k}$ & 237 & $310 \mathbf{k}$ & $-0.65$ \\
        \hline YAGO3-10 & $123 \mathbf{k}$ & 37 & $1 \mathbf{M}$ & $-0.54$ \\
        \hline
        \end{tabular}
    \label{tab:dataset}
\end{table}

\subsection{Inference  Patterns} 
UltraE can naturally infer relation patterns including symmetry, anti-symmetry, inversion and composition. As discussed above, the defined relation transformation $f_{r}=U_{\theta_r} B_{\mu_r} V_{\Phi_r}$ consists of three operations, including a circular rotation, a hyperbolic rotation, and a circular reflection.
The three operation matrices can all be identified as identity matrices.
Therefore, there are several different combinations of parameter settings to meet the above inferred requirements, demonstrating the comprehensive capability of the proposed UltraE on encoding relational patterns. 
For the sake of proof, we assume $B_{\mu_r}$ is an identity matrix $\mathbf{I}$, and $\Theta_r,\Phi_r\in[-\pi,\pi)$.
\begin{proposition}
Let $r$ be a symmetric relation such that for each triple $(e_h, r, e_t)$, its symmetric triple $(e_t, r, e_h)$ also holds. This symmetric property of $r$ can be encoded into UltraE.
\end{proposition}
\begin{proof}
If $r$ is a symmetric relation, by taking the $B_{\mathbf{b}_r}=\mathbf{I}$ and $U_{\Theta_r}=\mathbf{I}$, we have
\begin{align}
   \mathbf{e}_{h} = f_{r}\left(\mathbf{e}_{t}\right) =  V_{\Phi_r} \mathbf{e}_{t}, \ 
   \mathbf{e}_{t} = f_{r}\left(\mathbf{e}_{h}\right) =  V_{\Phi_r} \mathbf{e}_{h}
   \Rightarrow V_{\Phi_r}^2 = \mathbf{I} \nonumber
\end{align}
which holds true when $\Phi_r=0$ or $\Phi_r=-\pi$.
\end{proof}

\begin{proposition}
Let $r$ be an anti-symmetric relation such that for each triple $(e_h, r, e_t)$, its symmetric triple $(e_t, r, e_h)$ is not true. This anti-symmetric property of $r$ can be encoded into UltraE.
\end{proposition}
\begin{proof}
If $r$ is an anti-symmetric relation, by taking the $B_{\mathbf{b}_r}=\mathbf{I}$ and $U_{\Theta_r}=\mathbf{I}$, we have
\begin{align}
   \mathbf{e}_{h} = f_{r}\left(\mathbf{e}_{t}\right) =  V_{\Phi_r} \mathbf{e}_{t}, \ 
   \mathbf{e}_{t} = f_{r}\left(\mathbf{e}_{h}\right) =  V_{\Phi_r} \mathbf{e}_{h}
   \Rightarrow \mathbf{e}_{h}  = \mathbf{e}_{t} \nonumber
\end{align}
which holds true when $\Phi_r\neq0$ and $\Phi_r\neq-\pi$.
\end{proof}

\begin{proposition}
Let $r_1$ and $r_2$ be inverse relations such that for each triple $(e_h, r_1, e_t)$, its inverse triple $(e_t, r_2, e_h)$ is also true. This inverse property of $r_1$ and $r_2$ can be encoded into UltraE.
\end{proposition}
\begin{proof}
If $r_1$ and $r_2$ are inverse relations, by taking the $B_{\mathbf{b}_{r_1}}=B_{\mathbf{b}_{r_2}}=\mathbf{I}$ and $V_{\Phi_{r_1}}=V_{\Phi_{r_2}}=\mathbf{I}$, we have
\begin{align}
   \mathbf{e}_{h} = f_{r_1}\left(\mathbf{e}_{t}\right) =  U_{\Theta_{r_1}} \mathbf{e}_{t}, \ 
   \mathbf{e}_{t} = f_{r_2}\left(\mathbf{e}_{h}\right) =  U_{\Theta_{r_2}} \mathbf{e}_{h}
   \Rightarrow U_{\Theta_{r_1}}U_{\Theta_{r_2}} = \mathbf{I} \nonumber
\end{align}
which holds true when $\Theta_{r_1}+\Theta_{r_2}=0$.
\end{proof}

\begin{proposition}
Let relation $r_1$ be composed of $r_2$ and $r_3$ such that triple $(e_h, r_1, e_t)$ exists when $(e_h, r_2, e_t)$ and $(e_h, r_3, e_t)$ exist. This composition property can be encoded into UltraE.
\end{proposition}
\begin{proof}
If $r_1$ is composed of $r_2$ and $r_3$, by taking the $B_{\mathbf{b}_{r_1}}=B_{\mathbf{b}_{r_2}}=\mathbf{I}$ and $V_{\Phi_{r_1}}=V_{\Phi_{r_2}}=\mathbf{I}$, we have
\begin{align}
   &\mathbf{e}_{h} = f_{r_1}\left(\mathbf{e}_{t}\right) =  U_{\Theta_{r_1}} \mathbf{e}_{t}, \ 
   \mathbf{e}_{h} = f_{r_2}\left(\mathbf{e}_{t}\right) =  U_{\Theta_{r_2}} \mathbf{e}_{t}, \\
   &\mathbf{e}_{h} = f_{r_3}\left(\mathbf{e}_{t}\right) =  U_{\Theta_{r_3}} \mathbf{e}_{t} \
   \Rightarrow U_{\Theta_{r_1}} = U_{\Theta_{r_2}}U_{\Theta_{r_3}} \nonumber
\end{align}
which holds true when $\Theta_{r_1}=\Theta_{r_2}+\Theta_{r_3}$ or $\Theta_{r_1}=\Theta_{r_2}+\Theta_{r_3}+2\pi$ or $\Theta_{r_1}=\Theta_{r_2}+\Theta_{r_3}-2\pi$.
\end{proof}


\begin{table*}[]
    \centering
    \caption{Link prediction results (\%) on WN18RR, FB15k-237 and YAGO3-10 for low-dimensional embeddings ($d=32$) in the filtered setting. The first group of models are Euclidean models, the second groups are non-Euclidean models and MuRMP is a mixed-curvature baseline. RotatE, MuRE, MuRP, RotH, RefH and AttH results are taken from \citep{DBLP:conf/acl/ChamiWJSRR20}. 
    RotatE results are reported without self-adversarial negative sampling for fair comparison. The best score and best baseline are in \textbf{bold} and underlined, respectively. }
    \begin{tabular}{lrrrrrrrrrrrrrrr}
    \hline & \multicolumn{4}{c}{WN18RR} & \multicolumn{4}{c}{FB15k-237} & \multicolumn{4}{c}{ YAGO3-10} \\
    Model & MRR & H@ 1 & H@3 & H@10 & MRR & H@1 & H@3 & H@10 & MRR & H@1 & H@3 & H@10 \\
    \hline 
    TransE  & 36.6 & 27.4 & 43.3 & 51.5 & 29.5 & 21.0 & 32.2 & 46.6 & - & - & - & - \\
    RotatE  & 38.7 & 33.0 & 41.7 & 49.1 & 29.0 & 20.8 & 31.6 & 45.8 & - & - & - & - \\
    ComplEx & 42.1 & 39.1 & 43.4 & 47.6 & 28.7 & 20.3 & 31.6 & 45.6 & 33.6 & 25.9 & 36.7 & 48.4 \\
    QuatE   & 42.1 & 39.6 & 43.0 & 46.7 & 29.3 & 21.2 & 32.0 & 46.0 & - & - & - & - \\
    5$\star$E     & 44.9 & 41.8 & 46.2 & 51.0 & 32.3 & 24.0 & 35.5 & 50.1 & - & - & - & - \\
    MuRE    & 45.8 & 42.1 & 47.1 & 52.5 & 31.3 & 22.6 & 34.0 & 48.9 & 28.3 & 18.7 & 31.7 & 47.8 \\
    \hline 
    MuRP    & 46.5 & 42.0 & 48.4 & 54.4 & 32.3 & 23.5 & 35.3 & 50.1 & 23.0 & 15.0 & 24.7 & 39.2  \\
    RotH & \underline{47.2} & \underline{42.8} & \underline{49.0} & \underline{55.3} & 31.4 & 22.3 & 34.6 & 49.7 & 39.3 & 30.7 & 43.5 & 55.9 \\
    RefH & 44.7 & 40.8 & 46.4 & 51.8 & 31.2 & 22.4 & 34.2 & 48.9 & 38.1 & 30.2 & 41.5 & 53.0\\
    AttH & 46.6 & 41.9 & 48.4 & 55.1 & \underline{32.4} & \underline{23.6} & \underline{35.4} & 50.1 & \underline{39.7} & \underline{31.0} & \underline{43.7} & \underline{56.6} \\
    MuRMP & 47.0 & 42.6 & 48.3 & 54.7 & 31.9 & 23.2 & 35.1 & \underline{50.2} & 39.5 & 30.8 & 42.9 & \underline{56.6} \\
    \hline 
    UltraE (q=2) & 48.1 & 43.4 & 50.0 & 55.4 & 33.1 & 24.1 & 35.5 & 50.3 & 39.5 & 31.2 & 43.9 & 56.8  \\ 
    UltraE (q=4) & \textbf{48.8} & \textbf{44.0} & \textbf{50.3} & \textbf{55.8} & 33.4 & 24.3 & 36.0 & 51.0 & 40.0 & 31.5 & 44.3 & 57.0 \\ 
    UltraE (q=6) & 48.3 & 42.5 & 49.1 & 55.5 & \textbf{33.8} & \textbf{24.7} & \textbf{36.3} & \textbf{51.4} & \textbf{40.5} & \textbf{31.8} & \textbf{44.7} & \textbf{57.2} \\ 
    UltraE (q=8) & 47.5 & 42.3 & 49.0 & 55.1 & 32.6 & 24.6 & 36.2 & 51.0 & 39.4 & 31.3 & 43.4 & 56.5 \\ 
    \hline
\end{tabular}
\label{tab:low_dim}
\end{table*}

\section{Empirical Evaluation}
In this section, we evaluate the performance of UltraE on link prediction in three KGs that contain both hierarchical and non-hierarchical relations. 
We systematically study the major components of our framework and show that (1) UltraE outperforms Euclidean and non-Euclidean baselines on embedding KGs with heterogeneous topologies, especially in low-dimensional cases (Sec.~\ref{sec:low_dim}); (2) the signature of embedding space works as a knob for controlling the geometry, and hence influences the performance of UltraE (Sec.~\ref{sec:signature}); (3) UltraE is able to improve the embeddings of relations with heterogeneous topologies (Sec.~\ref{sec:topo}); and (4) the combination of rotation and reflection outperforms a single operator (Sec.~\ref{sec:operators}).

\subsection{Experiment Setup}

\subsubsection{Dataset.} We use three standard benchmarks: WN18RR \citep{DBLP:conf/nips/BordesUGWY13}, a subset of WordNet containing $11$ lexical relationships, FB15k-237 \citep{DBLP:conf/nips/BordesUGWY13}, a subset of Freebase containing general world knowledge, and YAGO3-10 \citep{DBLP:conf/cidr/MahdisoltaniBS15}, a subset of YAGO3 containing information of relationships between people. 
All three datasets contain hierarchical (e.g., \textit{partOf}) and non-hierarchical (e.g., \textit{similarTo}) relations, and some of which contain relational patterns like symmetry (e.g., \textit{isMarriedTo}). 
For each KG, we follow the standard data augmentation protocol \cite{DBLP:conf/icml/LacroixUO18} and the same train/valid/test splitting as used in \cite{DBLP:conf/icml/LacroixUO18} for fair comparision. 
Following the previous work \cite{DBLP:conf/acl/ChamiWJSRR20}, we use the global graph curvature \citep{DBLP:conf/iclr/GuSGR19} to measure the geometric properties of the datasets. The statistics of datasets are summarized in Table \ref{tab:dataset}. As we can see, all datasets are globally hierarchical (i.e., the curvature is negative) but none of which is a pure tree structure. 
Comparatively, WN18RR is more hierarchical than FB15k-237 and YAGO3-10 since it has a smaller global graph curvature.

\begin{table*}[]
    \centering
    \caption{Link prediction results (\%) on WN18RR, FB15k-237 and YAGO3-10 for high-dimensional embeddings (best for $d \in \{200,400,500\}$) in the filtered setting. RotatE, MuRE, MuRP, RotH, RefH and AttH results are taken from \citep{DBLP:conf/acl/ChamiWJSRR20}. RotatE results are reported without self-adversarial negative sampling. The best score and best baseline are in \textbf{bold} and underlined, respectively.}
    \begin{tabular}{lrrrrrrrrrrrrrr}
   \hline & \multicolumn{4}{c}{WN18RR} & \multicolumn{4}{c}{FB15k-237} & \multicolumn{4}{c}{ YAGO3-10} \\
    Model & MRR & H@ 1 & H@3 & H@10 & MRR & H@1 & H@3 & H@10 & MRR & H@1 & H@3 & H@10 \\
    \hline 
    TransE & 48.1 & 43.3 & 48.9 & 57.0 & 34.2 & 24.0 & 37.8 & 52.7 & - & - & - & - \\
    DistMult & 43.0 & 39.0 & 44.0 & 49.0 & 24.1 & 15.5 & 26.3 & 41.9 & 34.0 & 24.0 & 38.0 & 54.0 \\
    RotatE & 47.6 & 42.8 & 49.2 & 57.1 & 33.8 & 24.1 & 37.5 & 53.3 & 49.5 & 40.2 & 55.0 & 67.0 \\
   ComplEx & 48.0 & 43.5 & 49.5 & 57.2 & 35.7 & 26.4 & 39.2 & 54.7 & 56.9 & 49.8 & 60.9 & 70.1\\
    QuatE  & 48.8 & 43.8 & 50.8 & 58.2 & 34.8 & 24.8 & 38.2 & 55.0 & - & - & - & - \\
    5$\star$E  & \underline{50.0} & \underline{\textbf{45.0}} & 51.0 & \underline{59.0} & \underline{\textbf{37.0}} & \underline{\textbf{28.0}} & \underline{\textbf{40.0}} & 56.0 & - & - & - & -  \\
    MuRE & 47.5 & 43.6 & 48.7 & 55.4 & 33.6 & 24.5 & 37.0 & 52.1 & 53.2 & 44.4 & 58.4 & 69.4 \\
    \hline 
    MuRP & 48.1 & 44.0 & 49.5 & 56.6 & 33.5 & 24.3 & 36.7 & 51.8 & 35.4 & 24.9 & 40.0 & 56.7 \\
    RotH & 49.6 & 44.9 & \underline{51.4} & 58.6 & 34.4 & 24.6 & 38.0 & 53.5 & 57.0 & 49.5 & 61.2 & 70.6 \\
    RefH & 46.1 & 40.4 & 48.5 & 56.8 & 34.6 & 25.2 & 38.3 & 53.6 & \underline{57.6} & \underline{50.2} & \underline{61.9} & \underline{\textbf{71.1}} \\
    AttH & 48.6 & 44.3 & 49.9 & 57.3 & 34.8 & 25.2 & 38.4 & 54.0 & 56.8 & 49.3 & 61.2 & 70.2 \\
    MuRMP & 48.1 & 44.1 & 49.6 & 56.9 & 35.8 & 27.3 & 39.4 & \underline{56.1} & 49.5 & 44.8 & 59.1 & 69.8 \\
    \hline 
    UltraE (q=20) & 48.5 & 44.2 & 50.0 & 57.3 & 34.9 & 25.1 & 38.5 & 54.1 & 56.9 & 49.5 & 61.0 & 70.3 \\ 
    UltraE (q=40) & \textbf{50.1} & \textbf{45.0} & \textbf{51.5} & \textbf{59.2} & 35.1 & 27.5 & \textbf{40.0} & 56.0 & 57.5 & 49.8 & 62.0 & 70.8 \\ 
   UltraE (q=80) & 49.7 & 44.8 & 51.2 & 58.5 & 36.8 & 27.6 & \textbf{40.0} & \textbf{56.3} & \textbf{58.0} & \textbf{50.6} & \textbf{62.3} & \textbf{71.1} \\ 
   UltraE (q=160) & 48.6 & 44.5 & 50.3 & 57.4 & 35.4 & 26.0 & 39.0 & 55.5 & 57.0 & 49.5 & 61.8 & 70.5 \\ 
    \hline
\end{tabular}
\label{tab:high_dim}
\end{table*}

\noindent
\subsubsection{Evaluation protocol.} 
Two popular ranking-based metrics are reported: 1) mean reciprocal rank (MRR), the mean of the inverse of the true entity ranking in the prediction; and 2) hit rate $H@K$ ($K \in \{1,3,10\}$), the percentage of the correct entities appearing in the top $K$ ranked entities. As a standard, we report the metrics in the filtered setting \citep{DBLP:conf/nips/BordesUGWY13}, i.e., when calculating the ranking during evaluation, we filter out all true triples in the training set, since predicting a low rank for these triples should not be penalized.


\noindent
\subsubsection{Hyperparameters.} 
For each KG, we explore batch size $\in \{500,1000\}$, global margin $\in \{2,4,6,8\}$ and learning rate $\in \{3e-3,5e-3,7e-3\}$ in the validation set. The negative sampling size is fixed to $50$. The maximum number of epochs is set to $1000$. The radius of curvature $\alpha$ is fixed to $1$ since our model does not need relation-specific curvatures but is able to learn relation-specific mappings in the ultrahyperbolic manifold. The signature of the product manifold is set as the same as \citep{DBLP:conf/www/WangWSWNAXYC21}. 

\noindent
\subsubsection{Baselines.} Our baselines are divided into two groups:
\begin{itemize}
    \item \textbf{Euclidean models}. 1) TransE \citep{DBLP:conf/nips/BordesUGWY13}, the first translational model; 2) RotatE \citep{DBLP:conf/iclr/SunDNT19}, a rotation model in a complex space; 3) DistMult \citep{DBLP:journals/corr/YangYHGD14a}, a multiplicative model with a diagonal relational matrix; 4) ComplEx \citep{DBLP:conf/icml/TrouillonWRGB16}, an extension of DisMult in a complex space; 5) QuatE \citep{DBLP:conf/aaai/CaoX0CH21}, a generalization of complex KG embedding in a hypercomplex space ; 6) 5$\star$E that models a relation as five transformation functions; 7) MuRE \citep{DBLP:conf/nips/BalazevicAH19}, a Euclidean model with a diagonal relational matrix.
    \item \textbf{Non-Euclidean models.} 1) MuRP \citep{DBLP:conf/nips/BalazevicAH19}, a hyperbolic model with a diagonal relational matrix; 2) MuRS, a spherical analogy of MuRP; 3) RotH/RefH \citep{DBLP:conf/acl/ChamiWJSRR20}, a hyperbolic embedding with rotation or reflection; 4) AttH \citep{DBLP:conf/acl/ChamiWJSRR20}, a combination of RotH and RefH by attention mechanism; 5) MuRMP \citep{DBLP:conf/www/WangWSWNAXYC21}, a generalization of MuRP in the product manifold.
\end{itemize}
We compare UltraE with varied signatures (time dimensions). 
Since for all KGs, the hierarchies are much more dominant than cyclicity and we assume that both space and time dimension are even numbers, we set the time dimension to be a relatively small value (i.e., $q=2,4,6,8$ and $q=20,40,80,160$ for low-dimension settings and high-dimension settings, respectively) for comparison. A full possible setting of time dimension with $q \leq p$ is studied in Sec.~\ref{sec:signature}.

\subsection{Overall Results}
\label{sec:low_dim}
\subsubsection{Low-dimensional Embeddings.} 
Following previous non-Euclidean approaches \citep{DBLP:conf/acl/ChamiWJSRR20}, we first evaluate UltraE in the low dimensional setting ($d = 32$). Table \ref{tab:low_dim} shows the performance of UltraE and the baselines. Overall, it is clear that UltraE with varying time dimension  ($q=2,4,6,8$) improves the performance of all methods. UltraE, even with only $2$ time dimension, consistently outperforms all baselines, suggesting that the heterogeneous structure imposed by the pseudo-Riemannian geometry leads to better representations. 
In particular, the best performance of WN18RR is achieved by UltraE ($q=4$) while the best performances of FB15k-237 and YAGO3-10 are achieved by UltraE ($q=6$). We believe that this is because WN18RR is more hierarchical than FB15k-237 and YAGO3-10, validating our conjecture that the number of time dimensions controls the geometry of the embedding space. 
Besides, we observed that the mixed-curvature baseline MuRMP does not consistently improve the hyperbolic methods. We conjecture that this is because MuRMP cannot properly model relational patterns. 

\noindent
\subsubsection{High-dimensional Embeddings} 
Table \ref{tab:high_dim} shows the results of link prediction in high dimensions (best for $d \in \{200,400,500\}$). Overall, UltraE achieves either better or competitive results  against a variety of other models. 
In particular, we observed that there is no significant performance gain among hyperbolic methods and mixed-curvature methods against Euclidean-based methods. We conjecture that this is because when the dimension is sufficiently large, both Euclidean and hyperbolic geometries have sufficient ability to represent complex hierarchies in KGs. 
However, UltraE roughly outperforms all compared approaches, with the only exception of 5$\star$E achieving competitive results. Again, the performance gain is not as significant as in the low-dimension cases, which further validates the hypothesis that KG embeddings are not sensitive to the choice of embedding space with high dimensions. The additional performance gain might be obtained from the flexibility of inference of the relational patterns.

\begin{figure}
    \centering
    \includegraphics[width=0.45\textwidth]{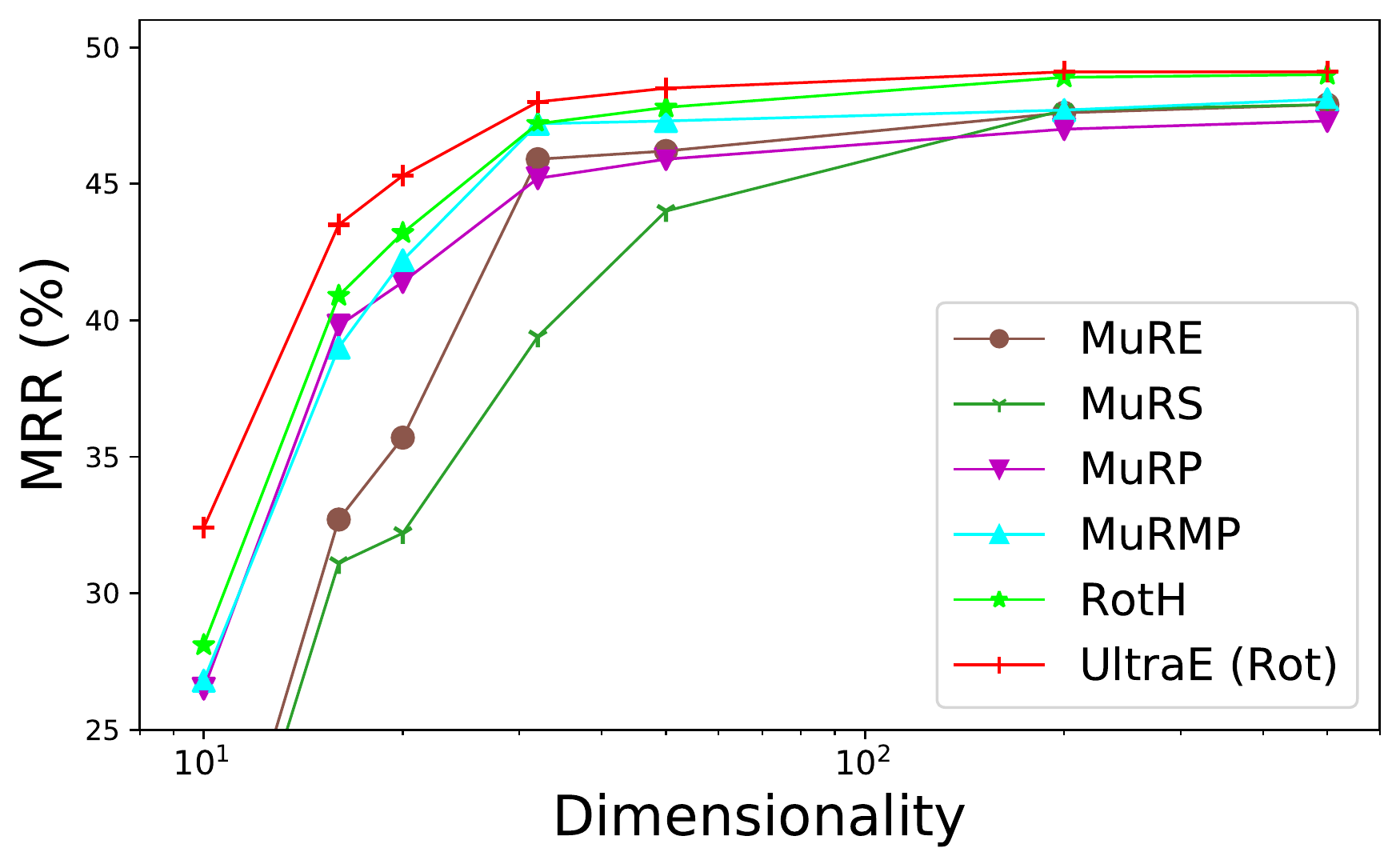}
    \caption{The performance (MRR) of various methods on WN18RR, with $d \in \{10,16,20,32,50,200,500\}$. UltraE is implemented with only rotation and $q=4$. 
    The results of MuRE, MuRS and MuRMP are taken from \cite{DBLP:conf/www/WangWSWNAXYC21} with $d \in \{10,15,20,40,100,200,500\}$. All results are averaged over 10 runs.}
    \label{fig:total_dim}
\end{figure}

\begin{figure}
    \centering
    \includegraphics[width=0.45\textwidth]{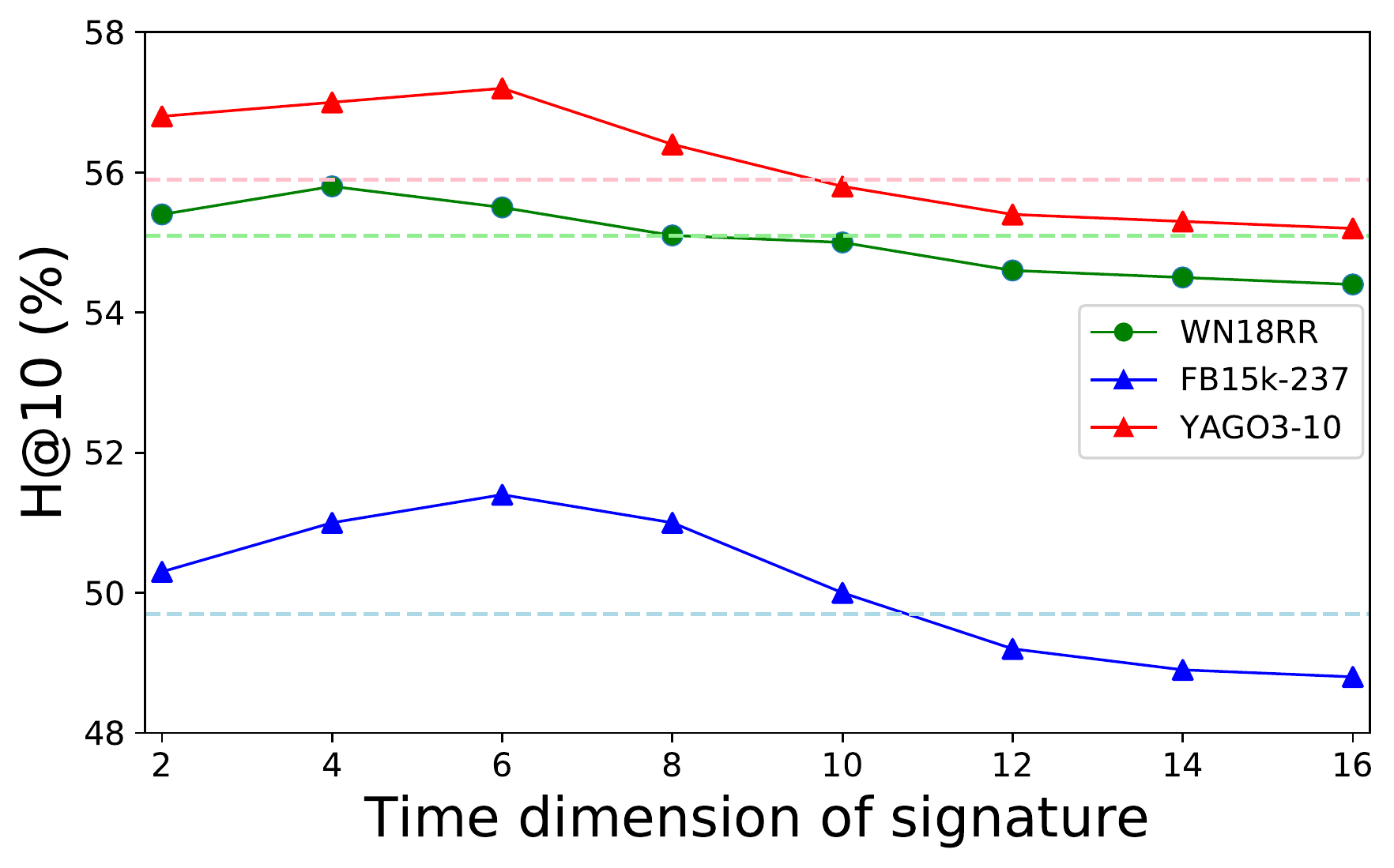}
    \caption{The performance (H@10) of UltraE with varied signature (time dimensions) under the condition of $d=p+q=32, q \leq p$ on WN18RR. The dashed horizontal lines denote the results of RotH. As $q$ increases, the performance first increases and starts to decrease after reaching a peak.}
    \label{fig:time_dim}
\end{figure}

\begin{table}[]
    \centering
    \vspace{0.1cm}
      \caption{Comparison of H@10 for WN18RR relations. Higher $\mathbf{Khs}_{G}$ and lower $\xi_{G}$ mean more hierarchical structure. UltraE is implemented by rotation and with best signature $(4,28)$. }
    \resizebox{\columnwidth}{!}{
    \begin{tabular}{lcccccc}
        \hline Relation & $\mathbf{Khs}_{G}$ & $\xi_{G}$ & RotE & RotH & UltraE (Rot)  \\
        \hline 
        \text {member meronym }              & 1.00 & -2.90 & 32.0 & 39.9 & 41.3   \\
        \text {hypernym }                    & 1.00 & -2.46 & 23.7 & 27.6 & 28.6   \\
        \text {has part }                    & 1.00 & -1.43 & 29.1 & 34.6 & 36.0  \\
        \text {instance hypernym }           & 1.00 & -0.82 & 48.8 & 52.0 & 53.2   \\
        \textbf {member of domain region }   & 1.00 & -0.78 & 38.5 & 36.5 & 43.3   \\
        \textbf { member of domain usage }    & 1.00 & -0.74 & 45.8 & 43.8 & 50.3  \\
        \text { synset domain topic of }      & 0.99 & -0.69 & 42.5 & 44.7 & 46.3   \\
        \text { also see }                    & 0.36 & -2.09 & 63.4 & 70.5 & 73.5   \\
        \hline
        \text { derivationally related form } & 0.07 & -3.84 & 96.0 & 96.8 & 97.1   \\
        \text { similar to }                  & 0.07 & -1.00 & 100.0 & 100.0 & 100.0  \\
        \text { verb group }                  & 0.07 & -0.50 & 97.4 & 97.4 & 98.0   \\
        \hline
    \end{tabular}}
    \label{tab:relation_type}
\end{table}

\begin{table}[]
    \centering
    \vspace{0.1cm}
     \caption{Comparison of H@10 on YAGO3-10 relations. UltraE (Rot) and UltraE (Ref) are implemented by only rotation and reflection, respectively. We choose the best signature $(6,26)$. }
    \resizebox{\columnwidth}{!}{
    \begin{tabular}{lccccc}
        \hline 
        Relation & Anti-symmetric & Symmetric & UltraE (Rot) & UltraE (Ref) & UltraE \\
        \hline 
        hasNeighbor & $\boldsymbol{x}$ & $\checkmark$ & 75.3 & \textbf{100.0} & \textbf{100.0} \\
        isMarriedTo & $\boldsymbol{x}$ & $\checkmark$ & 94.0 & 94.4 & \textbf{100.0} \\
        actedIn & $\checkmark$ & $\boldsymbol{x}$ & 14.7 & 12.7 & \textbf{15.3} \\
        hasMusicalRole & $\checkmark$ & $\boldsymbol{x}$ & 43.5 & 37.0 & \textbf{46.0} \\
        directed & $\checkmark$ & $\boldsymbol{x}$ & 51.5 & 45.3 & \textbf{56.8} \\
        graduatedFrom & $\checkmark$ & $\boldsymbol{x}$ & 26.8 & 16.3 & \textbf{27.5} \\
        playsFor & $\checkmark$ & $\boldsymbol{X}$ & 67.2 & 64.0 & \textbf{66.8} \\
        wroteMusicFor & $\boldsymbol{\checkmark}$ & $\boldsymbol{X}$ & \textbf{28.4} & 18.8 & 27.9  \\
        hasCapital & $\boldsymbol{\checkmark}$ & $\boldsymbol{X}$ & \textbf{73.2} & 68.3 & \textbf{73.2}  \\
        dealsWith & $\boldsymbol{X}$ &  $\boldsymbol{X}$ & 30.4 & 29.7 & \textbf{43.6} \\
        isLocatedIn & $\boldsymbol{X}$ & $\boldsymbol{X}$ & 41.5 & 39.8 & \textbf{42.8} \\
        \hline
    \end{tabular}}
    \label{tab:relational_pattern}
\end{table}

\subsection{Parameter Sensitivity}

\subsubsection{The effect of dimensionality.} 
To investigate the effect of dimensionality, we conduct experiments on WN18RR and compare UltraE ($q=4$) against various state-of-the-art counterparts with varying dimensionality. For a fair comparison with RotH that only considers rotation, we only use rotation for the implementation of UltraE, denoted by UltraE (Rot).
Fig. \ref{fig:total_dim} shows the results obtained by averaging over 10 runs. 
It clearly shows that the mixed-curvature method MuRMP outperforms its counterparts (MuRE, MuRP) with a single geometry, showcasing the limitation of a single homogeneous geometry on capturing the intrinsic heterogeneous structures. However, RotH performs slightly better than MuRMP, especially in high dimensionality, we conjecture that this is due to the capability of RotH on inferring relational patterns.  
UltraE achieves further improvements across a broad range of dimensions, suggesting the benefits of ultrahyperbolic manifold for modeling relation-specific geometries as well as inferring relational patterns.

\subsubsection{The effect of signature.}\label{sec:signature} We study the influence of the signature on WN18RR by setting a varying number of time dimensions under the condition of $d=p+q=32, p \geq q$. Fig. \ref{fig:time_dim} shows that in all three benchmarks, by increasing $q$, the performance grows first and starts to decline after reaching a peak, which is consistent with our hypothesis that the signature acts as a knob for controlling the geometric properties. One might also note that compared with hyperbolic baselines (the dashed horizontal lines), the performance gain for WN18RR is relatively smaller than those of FB15k-237 and YAGO3-10. We conjecture that this is because WN18RR is more hierarchical than FB15k-237 and YAGO3-10, and the hyperbolic embedding performs already well. This assumption is further validated by the fact that the best time dimension of WN18RR ($q=4$) is smaller than that of FB15k-237 and YAGO3-10 ($q=6$).

\subsubsection{The effect of relation types.}\label{sec:topo} In this part, we investigate the per-relationship performance of UltraE on WN18RR.
Similar to RotE and RotH that only consider rotation, we consider UltraE (Rot) as before. 
Two metrics that describe the geometric properties of each relation are reported, including global graph curvature and Krackhardt hierarchy score \citep{DBLP:conf/acl/ChamiWJSRR20}, for which higher $\mathbf{Khs}_{G}$ and lower $\xi_{G}$ means more hierarchical. 
As shown in Table \ref{tab:relation_type}, although RotH outperforms RotE on most of the relation types, the performance is not on par with RotE on relations "member of domain region" and "member of domain usage". UltraE (Rot), however, consistently outperforms both RotE and RotH on all relations, with significant performance gains on relations "member of domain region " and "member of domain usage " that RotH fails on. 
The overall observation also verifies the flexibility and effectiveness of the proposed method in dealing with heterogeneous topologies of KGs. 

\subsubsection{The effect of rotation and reflection.}\label{sec:operators}
To investigate the role of rotation and reflection, we compare UltraE against its two variants: UltraE with only rotation (UltraE (Rot)) and UltraE with only reflection (UltraE (Ref)). 
Table \ref{tab:relational_pattern} shows the per-relationship results on YAGO3-10. We observe that UltraE with rotation performs better on anti-symmetric relations while UltraE with reflection performs better on symmetric relations, suggesting that reflection is more suitable for representing symmetric patterns. On almost all relations including relations that are neither symmetric nor anti-symmetric, except for "wroteMusicFor", UltraE outperforms both rotation or reflection variants, showcasing that combining multiple operators can learn more expressive representations.

\section{Related Work}

\subsection{Knowledge Graph Embeddings}

Recent progress of KG embeddings has been achieved from many perspectives. One line of works aims at improving the expressivity of relational operations, from \emph{additive} operations \citep{DBLP:conf/nips/BordesUGWY13,DBLP:conf/aaai/WangZFC14,DBLP:conf/aaai/LinLSLZ15} to \emph{multiplicative} operations \citep{DBLP:conf/icml/NickelTK11,DBLP:journals/corr/YangYHGD14a,DBLP:conf/icml/LiuWY17}. Among which rotation model \citep{DBLP:conf/iclr/SunDNT19} allows for better representation of relational patterns such as symmetry, anti-symmetry, inversion and composition. 
Another line of work tries to exploit more expressive embedding space, from Euclidean space to hyperbolic space. Various hyperbolic KG embeddings are proposed, including MuRP \citep{DBLP:conf/nips/BalazevicAH19} that models relation as a combination of Möbius multiplication and Möbius addition, as well as RotH/RefH \citep{DBLP:conf/acl/ChamiWJSRR20} that models relations as hyperbolic isometries (rotation/reflection and translation) to infer relational patterns. RotH/RefH learn relation-specific curvature to distinguish the geometric characteristics of different relations, but still cannot tackle the non-hierarchical relations since hyperbolic space is not the optimal geometry of non-hierarchies. HyboNet \citep{DBLP:journals/corr/abs-2105-14686} is a multiplicative hyperbolic model in Lorentz geometry but requires a quadratic number of parameters.
5$\star$E \citep{DBLP:conf/aaai/NayyeriVA021} proposes $5$ transformations (inversion, reflection, translation, rotation, and homothety) to support multiple graph structures but the embeddings are still learned in the Euclidean space. Unlike all previous methods that focus on homogeneous geometric space, our method is learned in the ultrahyperbolic manifold, a heterogeneous geometric space with multiple kinds of local geometries. 

\subsection{Ultrahyperbolic Embeddings} 
Some recent works explored the application of ultrahyperbolic or pseudo-Riemannian geometry in representation learning. Pseudo-Riemannian geometry (or \textit{pseudo-Euclidean space}) was first applied to embed non-metric data that preserves local information \citep{sun2015space}. \citep{clough2017embedding} exploited Lorentzian space-time on embedding directed acyclic graphs. More recently, \cite{law2020ultrahyperbolic} proposed learning graph embeddings on pseudo-hyperboloid and provided some necessary geodesic tools, \cite{sim2021directed} further extended it into directed graph embedding, and \citep{DBLP:journals/corr/abs-2106-03134,law2021ultrahyperbolic} extended the pseudo-Riemannian embedding to support neural network operators. 
However, pseudo-Riemannian geometry has not yet been exploited in the setting of KG embeddings. 

\section{Conclusion}
This paper proposes UltraE, an ultrahyperbolic KG embedding method in a pseudo-Riemannian manifold that interleaves hyperbolic and spherical geometries, allowing for simultaneously modeling multiple hierarchical and non-hierarchical structures in KGs. 
We derive a relational embedding by exploiting the pseudo-orthogonal transformation, which is decomposed into various geometric operators including circular rotations/reflections and hyperbolic rotations, allowing for inferring complex relational patterns in KGs. We propose a Manhattan-like distance that measures the nearness of points in the ultrahyperbolic manifold.
The embeddings are optimized by standard gradient descent thanks to the differentiable and bijective mapping. 
We discuss theoretical connections of UltraE with other hyperbolic methods. 
On three standard KG datasets, UltraE outperforms many previous Euclidean and non-Euclidean counterparts, especially in  low-dimensional settings. 

\section*{Acknowledgments}
The authors thank the International Max Planck Research School for Intelligent Systems (IMPRS-IS) for supporting Bo Xiong. This project has received funding from the European Union’s Horizon 2020 research and innovation programme under the Marie Skłodowska-Curie grant agreement No: 860801.

\bibliographystyle{named}
\bibliography{main}



\end{document}